\documentclass[letterpaper]{article} 
\usepackage{aaai2026}  

\usepackage{times}  
\usepackage{helvet}  
\usepackage{courier}  
\usepackage[hyphens]{url}  
\usepackage{graphicx} 
\usepackage{natbib}  
\usepackage{caption} 
\usepackage{algorithm}
\usepackage{algorithmic}

\usepackage{newfloat}
\usepackage{listings}
\DeclareCaptionStyle{ruled}{labelfont=normalfont,labelsep=colon,strut=off} 
\floatstyle{ruled}
\newfloat{listing}{tb}{lst}{}
\floatname{listing}{Listing}
\usepackage{multicol}
\usepackage{multirow}
\usepackage{graphicx}
\usepackage{makecell}

\usepackage{comment}
\usepackage{subcaption}

\usepackage{times}
\usepackage{helvet}
\usepackage{courier}
\usepackage{xcolor}

\title{Detecting the Future: All-at-Once Event Sequence Forecasting with Horizon Matching} 

\author{Ivan Karpukhin\textsuperscript{\rm 1}\thanks{Corresponding author.}, 
Andrey Savchenko\textsuperscript{\rm 1,\rm 2,\rm 3} \\
}
\affiliations {
    \textsuperscript{\rm 1}Sber AI Lab, Moscow, Russia\\
    \textsuperscript{\rm 2}HSE University, Moscow, Russia \\
 \textsuperscript{\rm 3}ISP RAS Research Center for Trusted Artificial Intelligence, Moscow, Russia \\
    iakarpukhin@sberbank.ru, avsavchenko@hse.ru
}

\begin{document}

\maketitle

\begin{abstract}
Long-horizon events forecasting is a crucial task across various domains, including retail, finance, healthcare, and social networks. Traditional models for event sequences often extend to forecasting on a horizon using an autoregressive (recursive) multi-step strategy, which has limited effectiveness due to typical convergence to constant or repetitive outputs. To address this limitation, we introduce DEF, a novel approach for simultaneous forecasting of multiple future events on a horizon with high accuracy and diversity. Our method optimally aligns predictions with ground truth events during training by using a novel matching-based loss function.  We establish a new state-of-the-art in long-horizon event prediction, achieving up to a 50\% relative improvement over existing temporal point processes and event prediction models. 
Furthermore, we achieve state-of-the-art performance in next-event prediction tasks while demonstrating high computational efficiency during inference. 
\end{abstract}

\begin{links}
     \link{Code}{https://github.com/ivan-chai/hotpp-benchmark}
\end{links}

\section{Introduction}
Data from various domains, such as internet activity, e-commerce transactions, retail operations, and clinical visits, typically consists of timestamps with associated information. When ordered chronologically, these data points form event sequences, which differ fundamentally from other data types. Unlike tabular data \cite{wang2022transtab}, events inherently include timestamps and follow a specific temporal order. In contrast to time series data \cite{lim2021timeseriessurvey,ijcai2025p1266,savchenko2025hn}, event sequences exhibit irregular time intervals and often contain additional attributes. These unique characteristics necessitate the development of specialized models that can handle complex data streams. 
One of the primary tasks in the domain of event sequences is predicting future event types and their occurrence times \cite{xue2023easytpp,mcdermott2024eventstreamgpt}. It may be solved using the apparatus of Marked Temporal Point Processes (MTPP) \cite{rizoiu2017hawkes-tutorial} or their extensions for complex data streams that include additional event features~\cite{mcdermott2024eventstreamgpt}

Practical applications often require predicting multiple future events within a specified time horizon, such as forecasting purchases for the next month or making long-term medical prognoses \cite{xue2022hypro}. This task presents unique challenges that differ from traditional next-event prediction. The conventional approach typically relies on autoregressive models, which predict the next event step by step \cite{xue2023easytpp,xiao2018learninggan}. While these models are effective for immediate next-event forecasting, their performance tends to deteriorate as the prediction horizon extends \cite{karpukhin2024hotpp}. The same is true for horizon prediction models, including GAN \cite{xiao2018learninggan} and diffusion \cite{zhou2025diffusion}, which predict multiple future events at once but use pairwise losses between events on corresponding positions.

\begin{figure*}[t]
\begin{center}
  \includegraphics[width=0.82\textwidth]{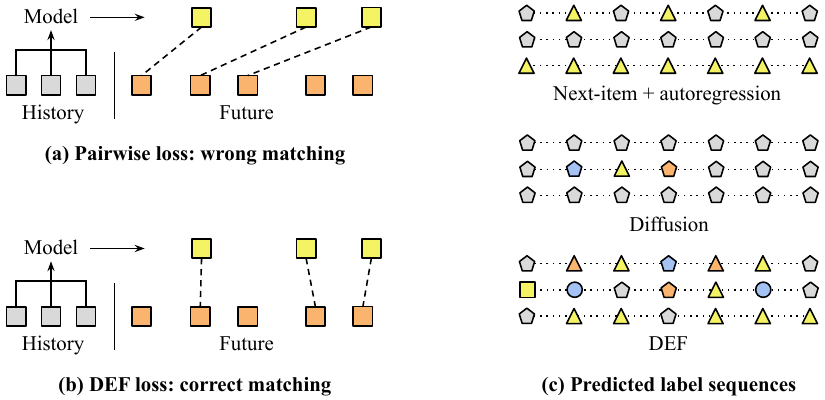}
\end{center}
\caption{(a) A typical next-event or pairwise loss compares events at corresponding positions, often resulting in incorrect matching. (b) The proposed matching loss calculates the loss function between the closest events, leading to a more robust and balanced error measure. (c) The proposed DEF method enhances the diversity of predictions. We demonstrate 3 example sequences from the Amazon dataset generated by the autoregressive IFTPP method, Diffusion, and the proposed approach. Each label type is depicted using a distinct shape and color combination. The precise timestamps are omitted for simplicity.}
\label{fig:motivation}
\end{figure*}

In this study, we identify significant limitations of pairwise losses in the context of long-horizon prediction. To address these challenges, we propose DEF (Detection-based Event Forecasting),
which detects multiple future events in parallel and employs a novel horizon matching loss, which dynamically aligns predictions with the closest ground-truth events, as illustrated in Fig.~\ref{fig:motivation}. This loss function enables the model to capture the full distribution of events within the horizon while remaining robust to outlier events. We demonstrate that our approach establishes a new state-of-the-art in long-horizon prediction, surpassing both autoregressive and horizon prediction approaches in terms of accuracy and prediction diversity. Additionally, our method exhibits high computational efficiency during inference, ranking among the fastest methods.

\section{Related Works}
\label{sec:related}
\paragraph{Event Sequences and Marked Temporal Point Processes.}
MTPP is a stochastic process that consists of a sequence of time-event pairs $(t_1, l_1), (t_2, l_2), \dots$, where $t_1 < t_2 < \dots$ denote the times of events, and $l_i \in \{1, \dots, L\}$ are the corresponding event type labels \cite{rizoiu2017hawkes-tutorial}. One of the most popular tasks in this domain is predicting the next event in the sequence~\cite{shchur2021neurtppreview,zhuzhel2023cotic}. A straightforward approach is to independently predict the time and type of the next event~\cite{shchur2019intensityfree,panos2024decomposable}, while more sophisticated methods model the temporal dynamics of each event type separately~\cite{mei2017nhp}. 
Traditional models, such as Poisson and Hawkes processes \cite{rizoiu2017hawkes-tutorial}, rely on strong assumptions about the underlying generative processes. Recent advancements have shifted towards more flexible and expressive models that leverage neural architectures. These include classical Recurrent Neural Networks (RNNs) \cite{du2016rmtpp,xiao2017rnnintencities,omi2019fullynn,shchur2019intensityfree},  as well as more advanced architectures like transformers \cite{zhang2020sahp,zuo2020thp,wang2022credittransformers,yang2022anhp,panos2024decomposable}. Additionally, continuous-time models such as Neural Hawkes Processes \cite{mei2017nhp}, ODE-RNN \cite{rubanova2019ode}, and their variants \cite{jia2019neuraljump,de2019gruodebayes,kidger2020neuralodebetter,song2024decoupledode,kuleshov2024cotode} have been developed to capture the dynamics of event sequences better. Moreover, generative models were adapted for use in MTPPs, including denoising diffusion~\cite{zhou2025diffusion,zeng2024interactingdiffusion} and Generative Adversarial Networks (GANs)~\cite{lin2022generativetpp}. Some prior works have explored general event sequences that include additional data fields beyond timestamps and labels \cite{mcdermott2024eventstreamgpt,padhi2021tabgpt}.

\paragraph{Long-horizon prediction.}
In the long-horizon event forecasting task, the goal is to predict future events within a time horizon $\mathrm{H}$, defined as the time interval $(t, t + \mathrm{H})$, where $t$ is the timestamp of the last observed event. A straightforward approach is to use autoregressive inference based on next-event prediction models. Previous research has also explored models that predict multiple future events simultaneously, known as next-$\mathrm{K}$ models \cite{karpukhin2024hotpp}, where $\mathrm{K}$ exceeds the typical number of events occurring within the horizon $\mathrm{H}$. These models are trained using pairwise losses that align predicted events with ground truth events at corresponding positions. Notably, next-event models can be viewed as a special case of next-1 models. Some approaches, such as GAN \cite{xiao2018learninggan} and Diffusion \cite{zhou2025diffusion,zeng2024interactingdiffusion}, incorporate pairwise losses as part of their training objectives. HYPRO \cite{xue2022hypro} has also addressed the problem of long-horizon prediction and introduced a technique for selecting the best candidate from a set of generated sequences. HYPRO functions as a meta-algorithm that can enhance the performance of nearly any sequence prediction model. However, its approach requires multiple generation runs for each prediction, significantly reducing training and inference efficiency.


Thus, previous studies have identified several challenges associated with autoregressive models for long-horizon predictions. 
These models often exhibit reduced prediction diversity and uncertainty over extended horizons, even though the task becomes increasingly complex. As illustrated in Fig.~\ref{fig:motivation}.c, the predicted label sequences often have constant or repetitive outputs. This behavior likely stems from the model's reliance on its predictions as input for subsequent predictions, which can amplify errors and lead to repeated events. Even horizon prediction approaches, such as Diffusion~\cite{zhou2025diffusion}, exhibit repetitive patterns because each of $\mathrm{K}$ outputs predicts the entire distribution of labels, leading to a bias toward the most frequent classes during inference. To address these issues, we propose a novel approach that offers greater diversity in its predictions while maintaining high accuracy for popular event forecasting tasks. 
The details are discussed in the following section.

\begin{figure*}
\begin{center}
  \includegraphics[width=.92\textwidth]{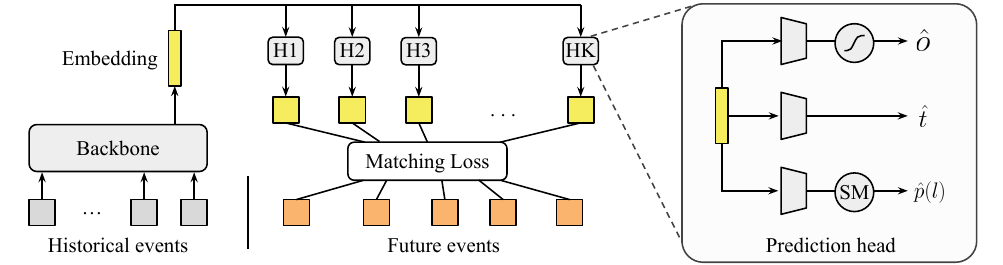}
  \caption{The proposed DEF simultaneously predicts $\mathrm{K}$ future events. Each prediction head outputs occurrence probability $\hat{o}$, time $\hat{t}$, and labels distribution $\hat{p}(l)$. During training, a novel matching loss aligns predictions with the ground truth sequence and evaluates its likelihood.}
  \label{fig:detpp}
\end{center}
\end{figure*}

\section{Proposed Approach}
\label{sec:detpp}
The proposed DEF method (Fig.~\ref{fig:detpp}) utilizes a backbone model to extract embeddings from historical data. Next, it simultaneously predicts $\mathrm{K}$ future event candidates by using multiple prediction heads. Here, the hyperparameter of our method, $\mathrm{K}$, should be larger than the typical sequence length in a specified time horizon $\mathrm{H}$. During training, the model aligns its predictions with ground truth and computes pairwise losses in the novel horizon matching loss function. At inference time, the model simultaneously predicts $\mathrm{K}$ events and retains only candidates with high prediction scores. Below, we provide a detailed overview of the event prediction head, the sequence model, and the associated training and inference procedures.

\paragraph{Prediction Head}
\label{sec:prediction-head}
Our method captures the complexity of event sequences by modeling each component of an event using a probabilistic framework that provides a rigorous basis for evaluating the likelihood of ground-truth event sequences.  Specifically, we propose predicting the probability of an event occurring, the distribution of event labels, and the distribution of time shifts relative to the last observed event. As depicted in Fig.~\ref{fig:detpp}, the probability $\hat{o}$ of an event occurring is modeled using a neural network with a sigmoid activation function. A separate head with softmax activation (SM) models the distribution $\hat{p}(l)$ of event labels. For the time shift, we use a Laplace distribution with a unit scale parameter, similarly to Mixture Density Networks (MDNs)~\cite{bishop1994mdn} and intensity-free MTPP~\cite{shchur2019intensityfree}:
\begin{equation}
    \mathrm{P}(t) = \frac{1}{2}e^{-|t - \hat{t}|},
\end{equation}
where $\hat{t}$ denotes the predicted time shift. This formulation offers a probabilistic interpretation of the MAE loss function. Note that a more rigorous formulation would involve using a truncated distribution to prevent negative time steps. However, preliminary experiments showed slightly worse performance with this approach, likely due to the reliance of most evaluation metrics on MAE.

By combining the predicted probabilities, we can estimate the likelihood of a future event given the output of the model:
\begin{equation}
\label{eq:head-prediction}
    \log \mathrm{P}(y) = \log \hat{o} + \log\hat{p}(l) -|t - \hat{t}| -\log \mathrm{R}(t).
\end{equation}
where $y = (t, l)$ represents an event with timestamp $t$ and label $l$. In Eq~\ref{eq:head-prediction}, we assume that, given the history of events, the timestamp, label, and occurrence of the next event are conditionally independent. The probability of a missed event (no event occurring) is given by:
\begin{equation}
    \log \mathrm{P}(\emptyset) = \log (1 -\hat{o}) + C_\emptyset,
\end{equation}
where $C_\emptyset$ is a constant independent of the model's output, representing the probability associated with a reserved ``unknown'' time and label values. To compute this loss, we omit $C_\emptyset$ since it does not influence the gradient during training.

\begin{table*}[t]
\centering
\begin{tabular}{lcccccccc}
Dataset & Domain & \thead{Sequences} & \thead{Events} & \thead{Avg. Length} & \thead{Classes} & \thead{Time\\unit} & \thead{OTD\\steps} & \thead{Horizon / \\ Mean length}
\\
\hline
StackOverflow & Social. net. & 2k & 138k & 64.2 & 22 & Minute & 10 & 10 / 12.0 \\
Amazon & Social. net. & 9k & 403k & 43.6 & 16 & N/A & 5 & 10 / 14.8 \\
Retweet & Social. net. & 23k & 1.3M & 56.4 & 3 & Second & 10 & 180 / 14.7 \\
MIMIC-IV & Medical & 120k & 2.4M & 19.7 & 34 & Day & 5 & 28 / 6.6 \\
Transactions & Financial & 50k & 43.7M & 875 & 203 & Day & 5 & 7 / 9.0 \\
\end{tabular}
\caption{Datasets statistics and evaluation parameters}
\label{tab:datasets}
\end{table*}

\paragraph{Horizon Matching Loss}
\label{sec:matching-loss}
Our approach is designed to predict $\mathrm{K}$ future events $\{\hat{y}_i\}_{i=1}^K$ within the time horizon $\mathrm{H}$. The set of ground truth events within this horizon is denoted by $\{y_i\}_{i=1}^T$, where the number of events $\mathrm{T}$ may vary. We propose to align the predicted sequence with the ground truth sequence by finding the matching that minimizes the following loss function, motivated by object detection techniques from computer vision, such as DeTR~\cite{carion2020detr}:
\begin{equation}
    \mathcal{L}(y, \hat{y})_{\mathrm{matching}} = \min\limits_{\sigma \in \mathcal{A}}\left[\sum\limits_{i=1}^{T}\mathcal{L}_\mathrm{pair}(y_i, \hat{y}_{\sigma(i)}) + \mathcal{L}_\mathrm{BCE}(\sigma, \hat{y})\right],
    \label{eq:matching}
\end{equation}
where $\mathcal{A}$ is the set of all possible alignments between the ground truth and predicted sequences and $\sigma$ represents a specific alignment. The optimal matching is computed using the Hungarian algorithm~\cite{kuhn1955hungarian}, which has a cubic computational complexity with respect to the sequence length. The pairwise loss $\mathcal{L}_\mathrm{pair}$ is similar to the negative log-likelihood of the ground truth event $y_i$ given the predicted distribution $\hat{y}_{\sigma(i)}$:
\begin{equation}
    \mathcal{L}_\mathrm{pair}(y_i, \hat{y}_{\sigma(i)}) = |t_i - \hat{t}_{\sigma(i)}| - \log \hat{p}_{\sigma(i)}(l_i),
\end{equation}
where $y = (t, l)$ is a ground truth event, $\hat{t}$ is the predicted timestamp, and $\hat{p}(l)$ is the predicted probability of the correct label.
The binary cross-entropy $\mathcal{L}_\mathrm{BCE}$ trains the model to predict the occurrence probability of events:
\begin{equation}
    \mathcal{L}_\mathrm{BCE}(\sigma, \hat{y}) = - \sum\limits_{i \in \sigma} \log{\hat{o_i}} - \sum\limits_{i \notin \sigma} \log(1 - \hat{o_i}),
\label{eq:bce}
\end{equation}
where $\hat{o}_i$ is the predicted probability that the $i$-th event is matched with some ground truth event.

By minimizing $\mathcal{L}_{\mathrm{matching}}(y, \hat{y})$, we train the model to accurately predict the parameters of the ground truth sequence, adapting to sequences of varying lengths, up to a maximum of $\mathrm{K}$ events. Unlike object detection training objectives such as DeTR \cite{carion2020detr}, DEF employs the same loss function for both the matching process and model training. Specifically, it integrates the alignment loss $\mathcal{L}_\mathrm{BCE}$ into the matching cost. 

To enable the method to address both the next-event prediction task and long-horizon forecasting accurately, we incorporate the next event prediction loss into the first output head. The final training objective is defined as follows:
\begin{equation}
    \mathcal{L}_\mathrm{DEF}(y, \hat{y}) = \mathcal{L}_{\mathrm{matching}}(y, \hat{y}) + \lambda\left[|t_1 - \hat{t}_1| - \log \hat{p}_1(l_1)\right].
    \label{eq:hybrid}
\end{equation}

\paragraph{Calibration and Inference}
Inference in our method involves two steps: filtering and sorting. First, predicted events are filtered based on their occurrence probabilities $\hat{o}_i$. The remaining events are sorted according to their predicted timestamps, forming the final output sequence. However, in practice, an additional calibration step is necessary. Without calibration, the model tends to predict a small number of events due to a bias in the predicted occurrence probabilities $\hat{o}_i$ toward the matching frequency of each head, which is typically below the 0.5 threshold. Calibration aims to determine optimal prediction thresholds for all \(\hat{o}_i\), aligning the prediction rates with the matching probabilities. This calibration is performed on the fly during training by tracking matching frequencies and computing the corresponding quantiles using a streaming algorithm. 
The calibration algorithm is outlined in the extended version of the paper. 


\paragraph{Conditional Head Architecture}
\label{sec:cond-head}
\begin{figure}
\begin{center}
  \includegraphics[width=\columnwidth]{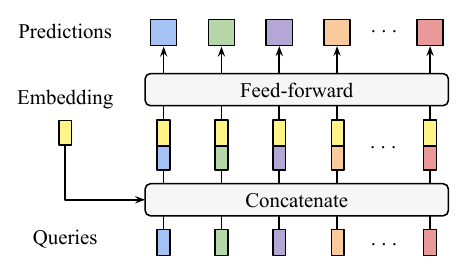}
\end{center}
\caption{Our conditional prediction head.}
    \label{fig:head}
\end{figure}
Implementing a separate feed-forward network for each prediction head leads to a large number of parameters and increases the risk of overfitting. Instead, we adopt a conditional approach that uses a single feed-forward network applied to $\mathrm{K}$ trainable query vectors, as illustrated in Fig.~\ref{fig:head}. The model receives two inputs: a query vector and a context vector, which are concatenated and passed through the shared feed-forward network. In this setup, each trainable query vector encodes the specific information required for its corresponding output head. This design enables the generation of all $\mathrm{K}$ outputs using a single network, significantly reducing the total number of parameters, accelerating convergence, and improving prediction quality. 

\paragraph{Hyperparameter Selection}
Our method has one key hyperparameter, the maximum number of predictions 
$\mathrm{K}$, which controls the maximum number of predictions. It typically requires tuning for each dataset, and we recommend setting $\mathrm{K}$ to approximately four times the average sequence length in a horizon for which it is required to predict future events. In our experiments, $\mathrm{K}$ values ranged from 32 to 64, depending on the characteristics of the dataset. 

Moreover, it may be necessary to assign the weights to each component of the proposed loss function. The value of $\lambda$ from Eq.~\ref{eq:hybrid} was set to 4 in all experiments. Additionally, we found it beneficial to adjust the weight of each loss component during alignment to accommodate the number of model outputs, dataset classes, and the average time step. In practice, the optimal weight for $\mathcal{L}_\mathrm{BCE}$ is typically around 8 times larger than the weights for the label and timestamp losses. For datasets with larger time steps, such as Retweet, the MAE loss weight should be reduced accordingly. The exact values of hyperparameters used in our experiments are provided in source code. 

\section{Experiments}
\label{sec:experiments}

\begin{table*}[t]
\begin{center}
\begin{tabular}{lccccc}
\multirow{2}{*}{Model} & \multicolumn{5}{c}{Metrics (OTD $\downarrow$ / T-mAP $\uparrow$)} \\
\cline{2-6}
& \multicolumn{1}{c}{StackOverflow} & \multicolumn{1}{c}{Amazon} & \multicolumn{1}{c}{Retweet} & \multicolumn{1}{c}{MIMIC-IV} & \multicolumn{1}{c}{Transactions} \\
\hline

\multirow{2}{*}{IFTPP} & 13.64 / 8.31\% & 6.52 / 22.56\% & 172.7 / 31.75\% & 11.53 / 21.67\% & 6.90 / 5.88\% \\
 & {\scriptsize  $\pm$0.05 / $\pm$0.50\%} & {\scriptsize  $\pm$0.05 / $\pm$0.52\%} & {\scriptsize  $\pm$4.4 / $\pm$4.44\%} & {\scriptsize  $\pm$0.01 / $\pm$0.21\%} & {\scriptsize  $\pm$0.01 / $\pm$0.13\%} \\
\hline
\multirow{2}{*}{IFTPP-T} & 13.61 / 8.81\% & 6.59 / 23.42\% & 166.1 / 40.05\% & {\bf 11.48} / 23.68\% & 6.85 / 5.55\% \\
 & {\scriptsize  $\pm$0.05 / $\pm$0.20\%} & {\scriptsize  $\pm$0.02 / $\pm$0.13\%} & {\scriptsize  $\pm$1.1 / $\pm$1.73\%} & {\scriptsize  $\pm$0.02 / $\pm$0.29\%} & {\scriptsize  $\pm$0.02 / $\pm$0.19\%} \\

\hline
\multirow{2}{*}{RMTPP} & 13.17 / 12.72\% & 6.57 / 20.06\% & 166.7 / 44.74\% & 13.71 / 21.08\% & 6.88 / 6.69\% \\
 & {\scriptsize  $\pm$0.05 / $\pm$0.16\%} & {\scriptsize  $\pm$0.03 / $\pm$0.33\%} & {\scriptsize  $\pm$3.3 / $\pm$0.94\%} & {\scriptsize  $\pm$0.03 / $\pm$0.29\%} & {\scriptsize  $\pm$0.01 / $\pm$0.12\%} \\
\hline
\multirow{2}{*}{NHP} & 13.24 / 11.96\% & 9.02 / 26.29\% & 165.8 / 45.07\% & 18.60 / 7.32\% & 6.98 / 5.61\% \\
 & {\scriptsize  $\pm$0.02 / $\pm$0.40\%} & {\scriptsize  $\pm$0.35 / $\pm$0.55\%} & {\scriptsize  $\pm$1.6 / $\pm$0.34\%} & {\scriptsize  $\pm$0.19 / $\pm$1.33\%} & {\scriptsize  $\pm$0.01 / $\pm$0.05\%} \\
\hline
\multirow{2}{*}{AttNHP} & 13.30 / 11.13\% & 7.30 / 14.62\% & 171.6 / 25.85\% & 14.68 / 22.46\% & 7.50 / 1.48\% \\
 & {\scriptsize  $\pm$0.02 / $\pm$0.32\%} & {\scriptsize  $\pm$0.06 / $\pm$0.80\%} & {\scriptsize  $\pm$1.0 / $\pm$1.08\%} & {\scriptsize  $\pm$0.08 / $\pm$0.40\%} & {\scriptsize  N/A / N/A} \\
\hline
\multirow{2}{*}{ODE} & 13.27 / 10.52\% & 9.46 / 22.96\% & 165.3 / 44.81\% & 14.74 / 15.18\% & 6.97 / 5.52\% \\
 & {\scriptsize  $\pm$0.03 / $\pm$0.23\%} & {\scriptsize  $\pm$0.08 / $\pm$0.61\%} & {\scriptsize  $\pm$0.5 / $\pm$0.69\%} & {\scriptsize  $\pm$0.34 / $\pm$0.15\%} & {\scriptsize  $\pm$0.01 / $\pm$0.13\%} \\
\hline
\multirow{2}{*}{HYPRO} & 13.26 / 14.69\% & 6.61 / 20.53\% & 170.7 / 46.99\% & 14.87 / 16.77\% & 7.05 / 7.05\% \\
 & {\scriptsize  N/A / N/A} & {\scriptsize  N/A / N/A} & {\scriptsize  N/A / N/A} & {\scriptsize  N/A / N/A} & {\scriptsize  N/A / N/A} \\
\hline
\multirow{2}{*}{Diffusion} & 13.01 / 15.07\% & 6.52 / 30.29\% & 158.0 / 52.24\% & 13.28 / 22.82\% & 6.88 / 6.04\% \\
 & {\scriptsize  $\pm$0.16 / $\pm$0.74\%} & {\scriptsize  $\pm$0.04 / $\pm$0.32\%} & {\scriptsize  $\pm$1.1 / $\pm$0.68\%} & {\scriptsize  $\pm$0.14 / $\pm$0.12\%} & {\scriptsize  $\pm$0.00 / $\pm$0.08\%} \\
\hline
 \multirow{2}{*}{\bf DEF} & {\bf 12.14} / {\bf 22.72\%} & { \bf 5.98 } / {\bf 37.20\%} & {\bf 132.9} / {\bf 57.93\%} & 12.95 / {\bf 30.35\%} & {\bf 6.70} / {\bf 9.26\%} \\
 & {\scriptsize  $\pm$0.04 / $\pm$0.32\%} & {\scriptsize  $\pm$0.04 / $\pm$0.06\%} & {\scriptsize  $\pm$0.7 / $\pm$0.33\%} & {\scriptsize  $\pm$0.32 / $\pm$0.25\%} & {\scriptsize  $\pm$0.03 / $\pm$0.09\%} \\
\hline
 \textit{Rel. Impr.} & $+6.7\%$ / $+50.8\%$ & $+8.3\%$ / $+22.8\%$ & $+15.9\%$ / $+10.9\%$ & $-12.8\%$ / $+28.2\%$ & $+2.2\%$ / $+31.3\%$ \\

\end{tabular}
\end{center}
\caption{Evaluation results in the long-horizon prediction task. The best result is shown in bold. Mean and STD values of 5 runs with different random seeds are reported. For HYPRO and AttNHP on the Transactions dataset, we report results using a single seed, as these methods exhibit low computational efficiency.}
\label{tab:horizon-results}
\end{table*}
\renewcommand{\arraystretch}{1}

We conducted a series of experiments using the HoTPP benchmark \cite{karpukhin2024hotpp} to assess the performance of our approach against several widely used MTPP models: IFTPP \cite{shchur2019intensityfree}, its transformer-based variant (IFTPP-T), intensity-based RMTPP \cite{du2016rmtpp} and NHP \cite{mei2017nhp} approaches, ODE-RNN \cite{rubanova2019ode}, the transformer-based AttNHP model \cite{yang2022anhp}, as well as the long-horizon HYPRO \cite{xue2022hypro} and Diffusion \cite{zhou2025diffusion} methods. Similar to IFTPP and RMTPP, DEF utilizes a Gated Recurrent Unit (GRU) \cite{cho2014gru} as its backbone. We employ the density variant of the IFTPP model and use a Laplace distribution, in place of a log-normal distribution, to better align with the target metrics.

The datasets employed in this study include Retweet \cite{zhao2015seismic}, Amazon \cite{jure2014amazon}, StackOverflow \cite{jure2014snap}, MIMIC-IV \cite{johnson2020mimic}, and Transactions \cite{agepred}, which represent a diverse range of domains and scales. Detailed dataset statistics are provided in Table~\ref{tab:datasets}.

MTPP models are typically evaluated based on their accuracy in predicting the next event \cite{xue2023easytpp}. Time and type predictions are often assessed separately, with type prediction quality measured by the error rate and time prediction evaluated using regression metrics such as Mean Absolute Error (MAE). Recent advancements have introduced additional metrics, including Optimal Transport Distance (OTD) \cite{mei2019imputing} and Temporal mAP (T-mAP) \cite{karpukhin2024hotpp}, which assess long-horizon predictions by comparing predicted sequences to ground truth sequences within a specified horizon. The horizon $\mathrm{H}$ should be chosen to align with the requirements of the specific task. In our experiments, we set $\mathrm{H}$ to match the horizon used in the T-mAP metric from the HoTPP benchmark~\cite{karpukhin2024hotpp}, ensuring consistency in evaluation.
The selected value is also sufficient to include the necessary number of events for calculating the OTD metric~\cite{mei2019imputing}. 
In this work, we employ all of the aforementioned metrics to evaluate the performance of our proposed method. Additionally, we measure prediction diversity using the entropy of the predicted labels distribution. 

Note that autoregressive methods typically condition on the ground-truth values of all preceding events when predicting the next event. This assumption becomes unrealistic for long-horizon forecasting. Consequently, we do not include log-likelihood evaluation in our experiments.

\begin{figure}
\begin{center}
  \includegraphics[width=\columnwidth]{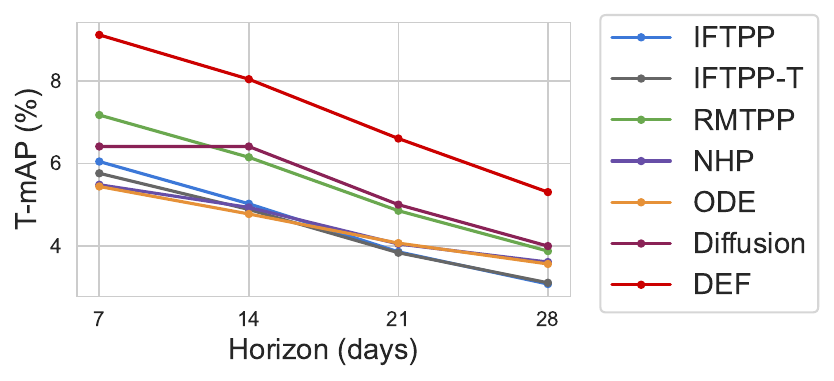}
\end{center}
\caption{Extended horizons prediction on the Transactions dataset.}
    \label{fig:longer-horizons}
\end{figure}

\begin{figure*}[h]
\begin{center}
  \includegraphics[width=\textwidth]{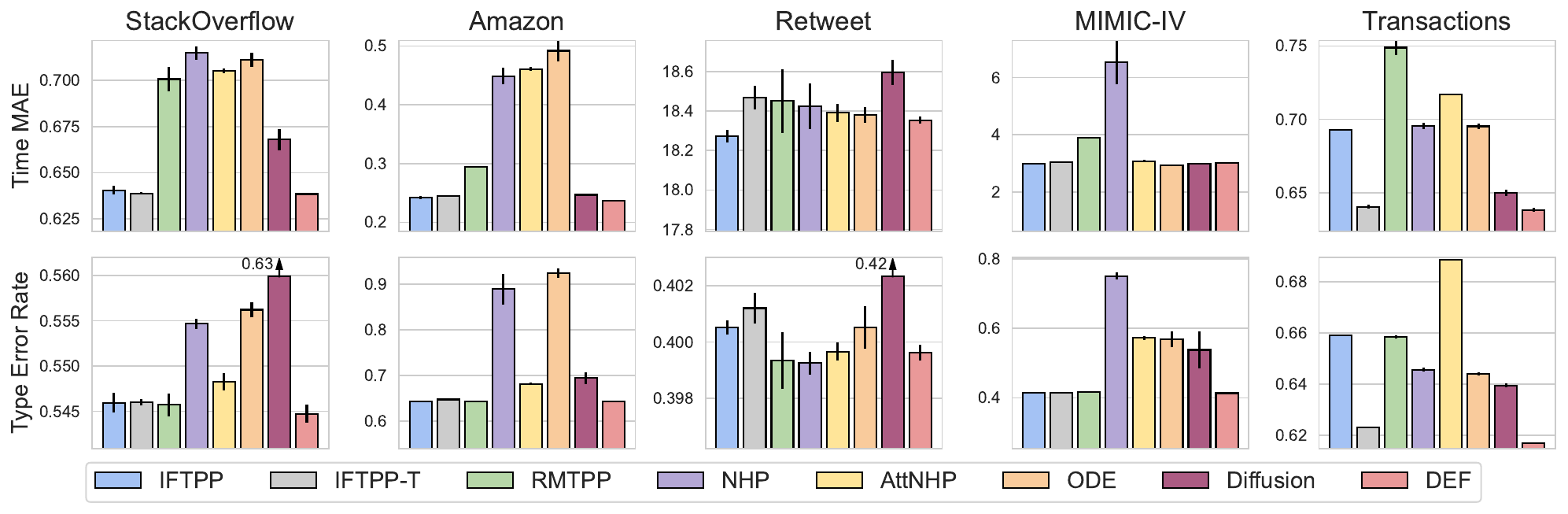}
  \caption{Next event prediction errors: MAE for time and error rate for type.}
  \label{fig:next-event}
\end{center}
\end{figure*}

\begin{figure*}[h]
\begin{center}
  \includegraphics[width=\textwidth]{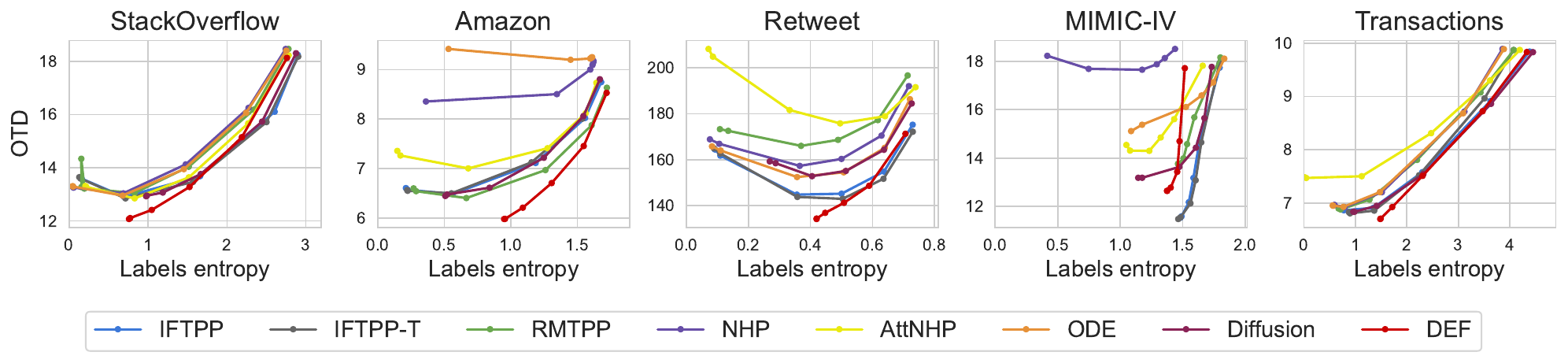}
  \caption{The relation between OTD and predictions' diversity for varying sampling temperature values. The optimal quality (low error and high diversity) corresponds to the bottom right corner.}
  \label{fig:entropy}
\end{center}
\end{figure*}

\begin{figure*}[h]
\begin{center}
  \includegraphics[width=\textwidth]{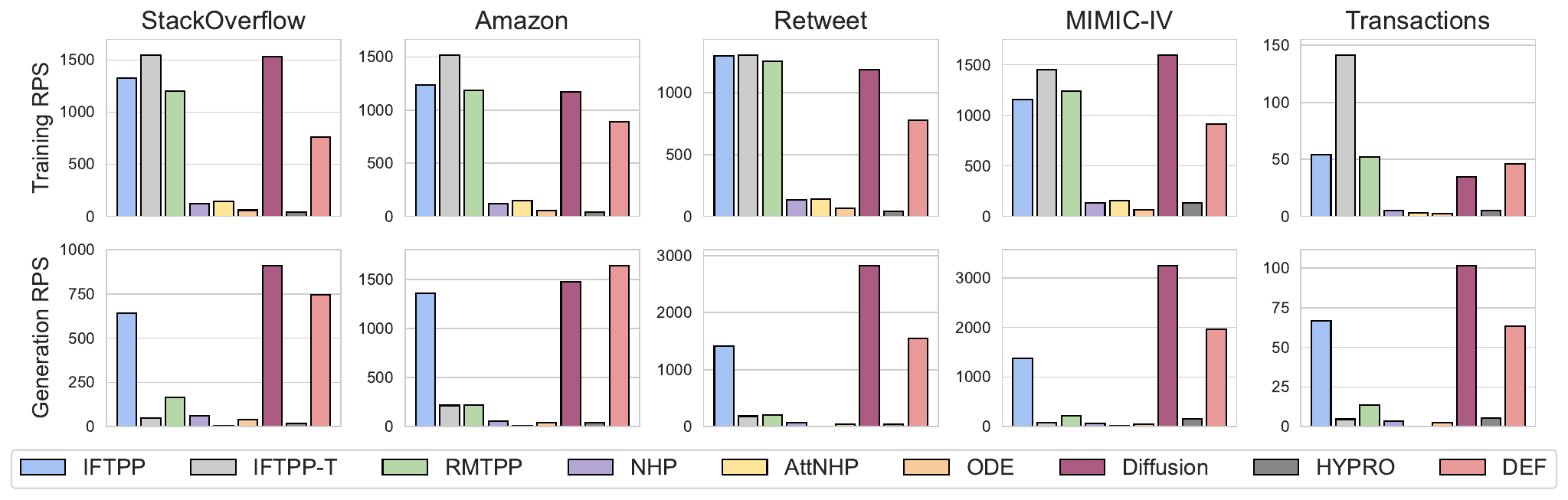}
  \caption{Training and sequence generation computation speed, Requests Per Second (RPS).}
  \label{fig:rps}
\end{center}
\end{figure*}

Further details on metric computation, ablation studies, additional experiments for general event sequences with multiple data fields, and the training process are provided in the extended version of the paper. 

\paragraph{Long-Horizon Events Forecasting}
\label{sec:horizon}

We evaluate long-horizon prediction performance using OTD and T-mAP. As shown in Table~\ref{tab:horizon-results}, DEF significantly outperforms existing approaches: it achieves state-of-the-art performance in 9 out of 10 comparisons. The only exception is the OTD metric on the MIMIC-IV dataset, where IFTPP ranks first and DEF second. 
The high T-mAP scores of our approach can be linked to its training objective, which utilizes matching, similar to T-mAP. However, it also consistently improves the OTD metric, suggesting that its training process enhances overall model performance rather than merely optimizing for a single evaluation criterion. 

While the base implementation of our method models a fixed prediction horizon $H$, this can be seen as a potential limitation. To address this, we explore a hybrid strategy that combines our approach with an autoregressive mechanism. Specifically, predictions over horizon $H$ are appended to the input sequence and used recursively to predict subsequent horizons. A similar autoregressive extension is applied to the Diffusion model for comparison. We evaluate this hybrid approach on the Transactions dataset, as its sequences are sufficiently long to support extended forecasting (see Table~\ref{tab:datasets} for dataset details). Results, shown in Figure~\ref{fig:longer-horizons}, demonstrate that the autoregressive variant of our method outperforms both traditional autoregressive baselines and fixed-horizon prediction models in capturing long-range event dynamics.

\paragraph{Next Event Prediction}
\label{sec:next-event}
Generative event sequences models and MTPPs are usually evaluated based on the quality of next-event prediction. We measured the next-event type error rate and mean absolute time error (MAE) across various datasets, with the results presented in Fig.~\ref{fig:next-event}. The proposed method achieves state-of-the-art results in all comparisons and significantly reduces error on the Transactions dataset, the only dataset where the difference between top methods is significant. Notably, the Transactions dataset also has the largest number of event types.
Thus, DEF, although primarily designed for long-horizon prediction, also achieves high-quality performance in the next-event prediction task.


\paragraph{Predictions Diversity}
\label{sec:diversity}
As demonstrated qualitatively in Fig.~\ref{fig:motivation}, popular autoregressive and horizon prediction methods often produce repetitive outputs. This section provides additional quantitative results to further emphasize the differences between traditional approaches and our method.

A common technique for increasing prediction diversity is adjusting the temperature during sampling. When the temperature approaches zero, the model selects the label with the highest probability, resulting in deterministic predictions. As the temperature increases, the model samples from a more uniform label distribution, leading to higher entropy and greater diversity. To analyze the relationship between temperature, prediction diversity, and long-horizon quality, we vary the temperature from 0 to 10, measuring the average entropy of predicted event types within the horizon. Additionally, we assess long-horizon prediction quality using OTD, as T-mAP is not affected by temperature changes.

Fig.~\ref{fig:entropy} indicates that our method achieves the best balance between diversity and prediction accuracy on four out of five datasets. While some methods benefit from sampling-based approaches compared to maximum-probability predictions, they still fail to match DEF’s overall performance. 



\paragraph{Training and Inference Speed}
\label{sec:speed}
A key practical consideration for any model is its computational efficiency. Fig.~\ref{fig:rps} compares the training and inference times of various methods in terms of Requests Per Second (RPS), i.e., the number of processed batch elements\footnote{Experiments were conducted on an Nvidia RTX 4060 GPU}. The results indicate that our approach exhibits a moderate training time while being one of the fastest methods during inference, alongside Diffusion and IFTPP.

The high inference speed of Diffusion is attributed to the low parameter count in its denoising model, optimized during hyperparameter tuning. In contrast, the proposed method employs a fully connected network for each output prediction, which impacts its RPS. Notably, all models utilizing NHP loss exhibit low training and inference speeds due to the computationally expensive sampling required at each step. Similarly, HYPRO demonstrates low computational efficiency, as it necessitates multiple autoregressive generations per step. Thus, our method ranks among the most computationally efficient methods in the field.

\section{Conclusion and Future Works}
In this work, we introduced a novel approach, called DEF, that addresses the challenges of long-horizon event forecasting by leveraging a matching-based training objective. Our experiments demonstrate that the proposed method effectively overcomes the limitations of traditional techniques, including specialized long-horizon models such as HYPRO and Diffusion. Notably, our approach not only achieves substantial improvements in long-horizon prediction accuracy but also sets a new state-of-the-art in next-event modeling, while generating more diverse predictions. Additionally, it offers greater computational efficiency at inference time by predicting multiple future events in parallel. This work advances event sequence modeling and opens new opportunities for a wide range of real-world applications.



We acknowledge several limitations of this work. First, we follow prior works~\cite{shchur2019intensityfree,du2016rmtpp} in assuming conditional independence between event attributes. While we focus primarily on the correct alignment of predictions and ground truth, the DEF architecture can be extended to model interdependencies. Second, similar to Diffusion~\cite{zhou2025diffusion}, we omit interdependencies between different events within the predicted horizon. Additional techniques, such as beam search or rescoring, could be incorporated to improve the model further. Finally, the training efficiency of our horizon matching loss relies heavily on the Hungarian algorithm, which may benefit from optimization of our custom CUDA kernel. 


Future research could investigate the integration of rescoring techniques, such as those in \cite{xue2022hypro}, or the application of beam search to enhance predictive performance. Additionally, better time modeling might be achieved by integrating intensity-based approaches, like NHP or RMTPP, with DEF, offering another promising direction for research. Furthermore, some techniques from our method could be adapted for object detection in computer vision \cite{carion2020detr}. We introduced a probabilistic framework that unifies different loss functions, using the same objective during matching and backpropagation, enhancing optimization robustness. We also employed an occurrence score during matching, which, as shown in our ablation studies, significantly improves the model performance on most datasets.

\section{Acknowledgments}
The work of A. Savchenko was supported by a grant, provided by the Ministry of Economic Development of the Russian Federation in accordance with the subsidy agreement (agreement identifier 000000C313925P4G0002) and the agreement with the Ivannikov Institute for System Programming of the Russian Academy of Sciences dated June 20, 2025 No. 139-15-2025-011.

\bibliography{main}



\end{document}